\documentclass[11pt,a4paper]{article}
\usepackage[hyperref]{emnlp2018}
\usepackage{times}

\usepackage[utf8]{inputenc}
\usepackage[T1]{fontenc}
\usepackage{amssymb}   %
\usepackage{amsmath}   %
\usepackage{amsfonts}  %
\usepackage{amsthm}    %
\usepackage{latexsym}  %
\usepackage{graphicx}  %
\usepackage{xcolor}    %
\usepackage{hyperref}  %
\usepackage{url}       %
\usepackage{xspace}    %
\usepackage{booktabs}  %
\usepackage{multirow}  %
\usepackage{multicol}  %
\usepackage{hhline}    %
\usepackage{ifthen}
\usepackage[noend]{algpseudocode}

\newcommand\refsec[1]{Section~\ref{sec:#1}}

\newcommand\reffig[1]{Figure~\ref{fig:#1}}

\newcommand\reftab[1]{Table~\ref{tab:#1}}

\ifthenelse{\isundefined{\definition}}{\newtheorem{definition}{Definition}}{}
\ifthenelse{\isundefined{\assumption}}{\newtheorem{assumption}{Assumption}}{}
\ifthenelse{\isundefined{\hypothesis}}{\newtheorem{hypothesis}{Hypothesis}}{}
\ifthenelse{\isundefined{\proposition}}{\newtheorem{proposition}{Proposition}}{}
\ifthenelse{\isundefined{\theorem}}{\newtheorem{theorem}{Theorem}}{}
\ifthenelse{\isundefined{\lemma}}{\newtheorem{lemma}{Lemma}}{}
\ifthenelse{\isundefined{\corollary}}{\newtheorem{corollary}{Corollary}}{}
\ifthenelse{\isundefined{\alg}}{\newtheorem{alg}{Algorithm}}{}
\ifthenelse{\isundefined{\example}}{\newtheorem{example}{Example}}{}

\aclfinalcopy %

\title{Mapping natural language commands to web elements}

\author{
  Panupong Pasupat \quad
  Tian-Shun Jiang \quad
  Evan Zheran Liu \quad
  Kelvin Guu \quad
  Percy Liang \\
  Stanford University \\
  {\tt \{ppasupat,jiangts,evanliu,kguu,pliang\}@cs.stanford.edu}
}

\date{}

\begin{document}
\maketitle

\begin{abstract}
The web provides a rich, open-domain environment with
textual, structural, and spatial properties.
We propose a new task for grounding language in this environment:
given a natural language command
(e.g., ``click on the second article''),
choose the correct element on the web page (e.g., a hyperlink or text box).
We collected a dataset of over 50,000 commands that
capture various phenomena such as
functional references (e.g. ``find who made this site''),
relational reasoning (e.g. ``article by john''),
and visual reasoning (e.g. ``topmost article'').
We also implemented and analyzed three baseline models
that capture different phenomena
present in the dataset. \end{abstract}

\section{Introduction}

Web pages are complex documents containing both structured properties
(e.g., the internal tree representation) %
and unstructured properties
(e.g., text and images).
Due to their diversity in content and design,
web pages provide a rich environment
for natural language grounding tasks.

\begin{figure}[t]\centering
\includegraphics[width=\columnwidth]{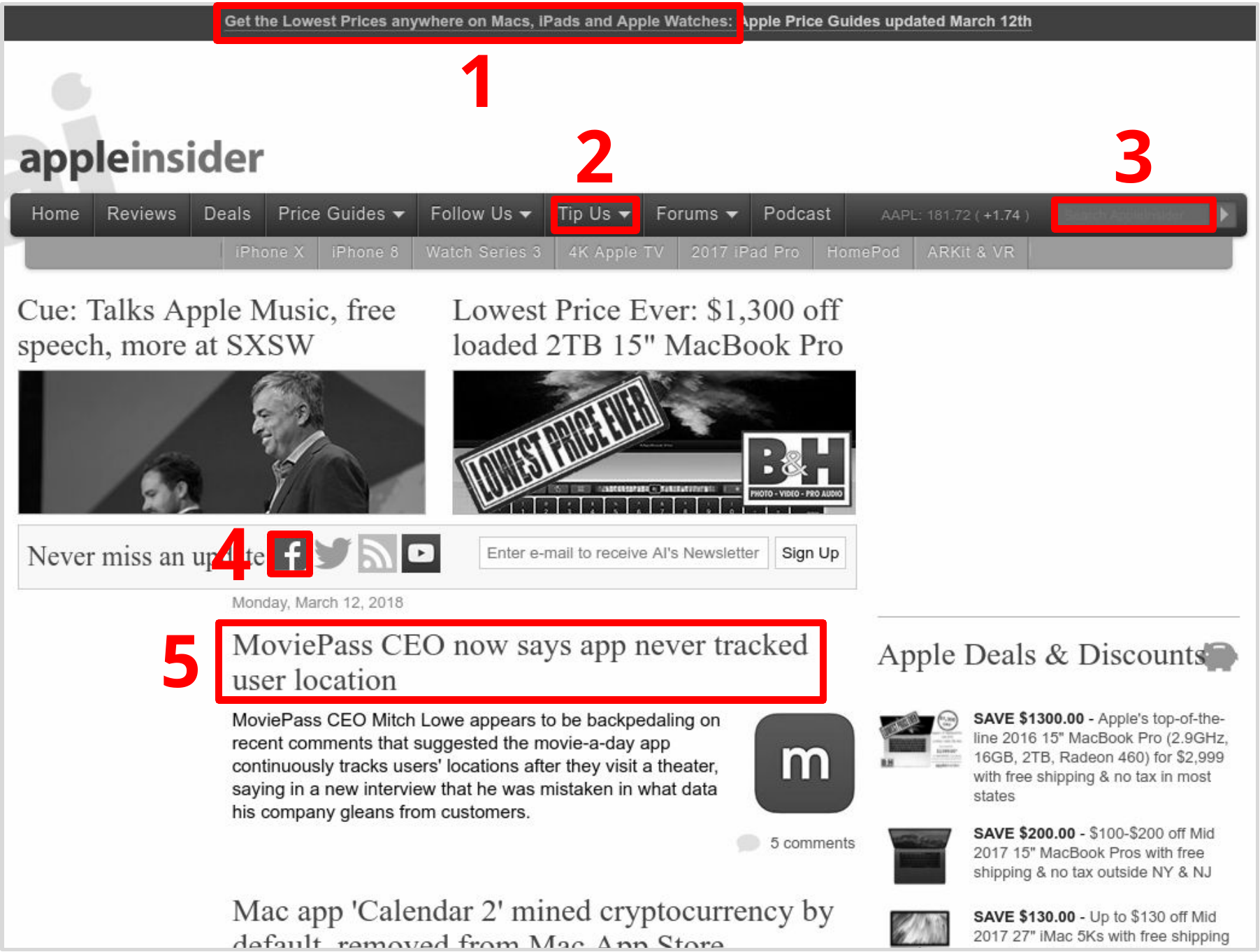}\\[.5em]
{\small
\begin{tabular}{@{}r@{\;}lr@{\;}l@{}}
1: & click on apple deals &
2: & send them a tip \\
3: & enter iphone 7 into search &
4: & follow on facebook \\
5: & open most recent news update \\
\end{tabular}
}
\caption{
Examples of natural language commands on
the web page \texttt{appleinsider.com}.
}\label{fig:running-example}
\end{figure}

In particular, we consider the task of mapping natural language commands
to web page elements (e.g., links, buttons, and form inputs),
as illustrated in \reffig{running-example}.
While some commands refer to an element's text directly,
many others require more complex reasoning with the various aspects of web pages:
the text, attributes, styles,
structural data from the document object model (DOM),
and spatial data from the rendered web page.

Our task is inspired by the
semantic parsing literature,
which aims to map natural language utterances
into actions such as database queries and object manipulation
\cite{zelle96geoquery,chen11navigate,artzi2013weakly,berant2013freebase,%
misra2015environment,andreas2015alignment}.
While these actions usually act on an environment
with a fixed and known schema,
web pages contain a larger variety of structures,
making the task more open-ended.
At the same time, our task can be viewed as a reference game
\cite{golland2010pragmatics,smith2013pragmatics,andreas2016reasoning},
where the system has to select an object given a natural language reference.
The diversity of attributes in web page elements,
along with the need to use context to interpret elements,
makes web pages particularly interesting.

Identifying elements via natural language
has several real-world applications.
The main one is providing a voice interface for interacting with web
pages,
which is especially useful as an assistive technology for the visually impaired
\cite{zajicek1998blind,ashok2014wizard}.
Another use case is browser automation:
natural language commands are less brittle than CSS or XPath selectors
\cite{hammoudi2016why}
and could generalize across different
websites.

\begin{table*}[t]\centering\small
\bgroup\def\arraystretch{1.2}
\begin{tabular}{|l|p{5.3cm}|p{5.5cm}|r|}\hline
\textbf{Phenomenon} & \textbf{Description} & \textbf{Example} & \textbf{Amount} \\ \hline
substring match &
The command contains only a substring of the element's text (after stemming). &
``view internships with energy.gov'' $\to$ ``Careers \& Internship'' link &
7.0 \% \\ \hline
paraphrase &
The command paraphrases the element's text. &
``click sign in'' $\to$ ``Login'' link &
15.5 \% \\ \hline
goal description &
The command describes an action or asks a question. &
``change language'' $\to$ a clickable box with text ``English'' &
18.0 \% \\ \hline
summarization &
The command summarizes the text in the element. &
``go to the article about the bengals trade''
$\to$ the article title link &
1.5 \% \\ \hline
element description &
The command describes a property of the element. &
``click blue button'' &
2.0 \% \\ \hline
relational reasoning &
The command requires reasoning with another element or its surrounding context. &
``show cookies info''
$\to$ ``More Info'' in the cookies warning bar, not in the news section &
2.5 \% \\ \hline
  ordinal reasoning &
The command uses an ordinal. &
``click on the first article'' &
3.5 \% \\ \hline
spatial reasoning &
The command describes the element's position. &
``click the three slashes at the top left of the page'' &
2.0 \% \\ \hline
image target &
The target is an image (no text). &
``select the favorites button'' &
11.5 \% \\ \hline
form input target &
The target is an input (text box, check box, drop-down list, etc.). &
``in the search bar, type testing'' &
6.5 \% \\ \hline
\end{tabular}
\egroup
\caption{
Phenomena present in the commands in the dataset.
Each example can have multiple phenomena.
}\label{tab:analysis}
\end{table*}
 
We collected a dataset of over 50,000 natural language commands.
As seen in \reffig{running-example},
the commands contain many phenomena,
such as relational, visual, and functional reasoning,
which we analyze in greater depth in \refsec{dataset-analysis}.
We also implemented three models for the task
based on retrieval, element embedding, and text alignment.
Our experimental analysis shows that
functional references, relational references, and visual reasoning
are important for correctly identifying elements from natural language
commands.
 
\section{Task}

Given a web page $w$ with elements $e_1, \dots, e_k$ and a command $c$,
the task is to select the element $e \in \{e_1,\dots,e_k\}$ described by the command $c$.
The training and test data contain $(w, c, e)$ triples.

\subsection{Dataset}

We collected a dataset of 51,663 commands on 1,835 web pages.
To collect the data, we first archived home pages of the top 10,000
websites\footnote{\scriptsize\url{https://majestic.com/reports/majestic-million}}
by rendering them in Google Chrome.
After loading the dynamic content,
we recorded the DOM trees and the geometry of each element,
and stored the rendered web pages.
We filtered for web pages in English that rendered correctly
and did not contain inappropriate content.
Then we asked crowdworkers to brainstorm different actions for each web page,
requiring each action to reference exactly one element (of their choice) from the
filtered list of interactive elements (which include
visible links, inputs, and buttons).
We encouraged the workers to avoid using the exact words of the
elements by granting a bonus for each command that did not contain the exact wording
of the selected element.
Finally, we split the data into 70\% training, 10\% development, and 20\% test data.
Web pages in the three sets do not overlap.

The collected web pages have an average of 1,051 elements,
while the commands are 4.1 tokens long on average.

\subsection{Phenomena present in the commands}\label{sec:dataset-analysis}

Apart from referring to the exact text of the element,
commands can refer to elements in a variety of ways.
We analyzed 200 examples from the training data and broke down the phenomena present in these commands (see \reftab{analysis}).

Even when the command directly references the element's text,
many other elements on the page also have word overlap with the command.
On average, commands have word overlap with 5.9 leaf elements on the page (not counting stop words).

\section{Models}

\subsection{Retrieval-based model}

Many commands refer to the elements by their text contents.
As such,
we first consider a simple retrieval-based model that
uses the command as a search query to retrieve the most relevant element
based on its TF-IDF score.

Specifically, each element is represented as a bag-of-tokens computed by
(1) tokenizing and stemming its text content,
and (2) tokenizing the attributes (id, class, 
placeholder, label, tooltip, aria-text, name, src, href)
at punctuation marks and camel-case boundaries.
When computing term frequencies,
we downweight the attribute tokens from (2) by a factor of $\alpha$.
We use $\alpha = 3$ tuned on the development set
for our experiments.

The document frequencies are computed over the web pages in the training dataset.
If multiple elements have the same score,
we heuristically pick the most prominent element, i.e.,
the one that appears earliest in the pre-order traversal of the DOM hierarchy.

\subsection{Embedding-based model}

A common method for matching two pieces of text
is to embed them separately and then compute a score from the two embeddings
\cite{kiros2015skip,tai2015improved}.
For a command $c$ and elements $e_1, \dots, e_k$,
we define the following conditional distribution over the elements:
\[p\,(e_i\mid c) \propto \exp\left[s(f(c), g(e_i))\right]\]
where $s$ is a scoring function,
$f(c)$ is the embedding of $c$,
and $g(e_i)$ is the embedding of $e_i$,
described below.
The model is trained to maximize the log-likelihood of the correct element in the training data.

\paragraph{Command embedding.}
To compute $f(c)$,
we embed each token of $c$ into a fixed-dimensional vector
and take an average\footnote{We tried applying LSTM but found no improvement.} over the token embeddings.
(The token embeddings are initialized with GloVe vectors.)

\begin{figure}[t]\centering
\includegraphics[scale=.4]{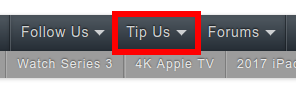}\\
{\small
\texttt{<a class="dd-head" id="tip-link" href="submit\_story/">Tip Us</a>}\\[.8em]
\begin{tabular}{rl}
Text content: & tip us \\
String attributes: & tip link dd head \\
Visual features: & location = (0.53, 0.08) \\ & visible = true
\end{tabular}
}
\caption{
Example of properties used to compute
the embedding $g(e)$ of the element $e$.
}\label{fig:features}
\end{figure}

\paragraph{Element embedding.}
To compute $g(e)$,
we embed the properties of $e$,
concatenate the results, and then apply a linear layer
to obtain a vector of the same length as $f(c)$.
\reffig{features} shows an example of the properties
that the model receives.
The properties include:
\begin{itemize}
\item \emph{Text content.}
We apply the command embedder $f$ on the text content of $e$.
As the text of most elements of interest (links, buttons, and inputs)
are short, we find it sufficient to limit the text to the first 10 tokens
to save memory.
\item \emph{Text attributes.}
Several attributes
(aria, title, tooltip, placeholder, label, name)
usually contain natural language.
We concatenate their values and then apply the command embedder $f$
on the resulting string.
\item \emph{String attributes.}
We tokenize other string attributes
(tag, id, class) at punctuation marks and camel-case boundaries.
Then we embed them with separate lookup tables and average the resulting vectors.
\item \emph{Visual features.}
We form a vector consisting of the coordinates of the element's center
(as fractions of the page width and height)
and visibility (as a boolean).
\end{itemize}

\paragraph{Scoring function.}
To compute the score $s(f(c), g(e))$,
we first let $\hat{f}(c)$ and $\hat{g}(e)$ be the results of normalizing the two embeddings to unit norm.
Then we apply a linear layer on the concatenated vector
$[\hat{f}(c); \hat{g}(e); \hat{f}(c)\circ \hat{g}(e)]$
(where $\circ$ denotes the element-wise product).

\paragraph{Incorporating spatial context.}
Context is critical in certain cases;
for example, selecting a text box relies on knowing
the neighboring label text,
and selecting an article based on the author
requires locating the author's name nearby.
Identifying which related element should be considered %
based on the command is a challenging task.

We experiment with adding \emph{spatial context}
to the model.
For each direction $d \in$ \{top, bottom, left, right\},
we use $g$ to embed a neighboring element $n_d(e)$ directly adjacent to $e$ in that direction.
(If there are multiple such elements, sample one;
if there is no such element, use a zero vector.)
After normalizing the results to get $\hat{g}(n_d(e))$,
we concatenate $\hat{g}(n_d(e))$ and $\hat{f}(c)\circ \hat{g}(n_d(e))$
to the linear layer input in
the scoring function.

\subsection{Alignment-based model}

One downside of the embedding-based model is that
the text tokens from $c$ and $e$ do not directly interact.
Previous works on sentence matching
usually employ either unidirectional or
bidirectional attention to tackle this issue
\cite{seo2016bidaf,yin2016abcnn,xiong2017dynamic,yu2018qanet}.
We opt for a simple method based on a single alignment matrix similar to
\citet{hu2014convolutional}
as described below.

Let $t(e)$ be the concatenation of $e$'s text content and text attributes of $e$,
trimmed to 10 tokens.
We construct a matrix $A(c,e)$ where each entry $A_{ij}(c,e)$
is the dot product between the embeddings of the $i$th token of $c$
and the $j$th token of $t(e)$.
Then we apply two convolutional layers of size $3 \times 3$ on the matrix,
apply a max-pooling layer of size $2\times 2$, concatenate a tag embedding,
and then apply a linear layer on the result to get a 10-dimensional vector $h(c,e)$.

We apply a final linear layer on $h(c,e)$ to compute a scalar score,
and then train on the same objective function as the encoding-based model.
To incorporate context, we simply concatenate the four vectors $h(c,n_d(e))$
of the neighbors $n_d(e)$ to the final linear layer input.
 
\section{Experiments}

We evaluate the models on
\emph{accuracy}, the fraction of examples that the model selects the correct element.
We train the neural models
using Adam \cite{kingma2014adam}
with initial learning rate $10^{-3}$,
and apply early stopping based on the development set.
The models can choose any element that is visible on the page at rendering time.

The experimental results are shown in \reftab{main-results}.
Both neural models significantly outperform the retrieval model.
\subsection{Ablation analysis}

To measure the importance of each type of information in web pages,
we perform an ablation study where the model does not observe
one of the following aspects of the elements:
text contents, attributes, and spatial context.

Unsurprisingly, the results in \reftab{main-results}
show that text contents are the most important input signal.
However, attributes also play an important role
in both the embedding and alignment models.
Finally, while spatial context increases
alignment model performance, the gain is very small, suggesting that
incorporating appropriate contexts to the model is a challenging task
due to the variety in the types of context,
as well as the sparsity of the signals.

\begin{table}[t]\centering
\small
\begin{tabular}{|l|c|c|}\hline
\multicolumn{1}{|c|}{\textbf{Model}} & \textbf{Accuracy (\%)} \\ \hline %
retrieval & 36.55 \\ %
\hline
embedding & 56.05 \\ %
\hspace*{.5em} no texts & 23.62 \\ %
\hspace*{.5em} no attributes & 55.43 \\ %
\hspace*{.5em} no spatial context & 58.87 \\ %
\hline
alignment & 50.74 \\ %
\hspace*{.5em} no texts & 15.94 \\ %
\hspace*{.5em} no attributes & 48.51 \\ %
\hspace*{.5em} no spatial context  & 50.66 \\ %
\hline
\end{tabular}
\caption{
Accuracies of the models and their ablations.
}\label{tab:main-results}
\end{table}

\subsection{Error analysis}
To get a better picture of how the models handle different phenomena,
we analyze the predictions of the embedding-based and alignment-based models
on 100 development examples
where at least one model made an error.
The errors, summarized in \reftab{errors},
are elaborated below:

\begin{table}[t]\centering
\small
\begin{tabular}{|l|r|r|}\hline
\textbf{Error Type} & \textbf{Embed} & \textbf{Align} \\
\hline
Fail to match strings & 26.8\% & 11.6\% \\
Incorrectly match strings & 3.8\% & 14.2\% \\
Fail to understand paraphrases & 8.9\% & 7.9\% \\
Fail to understand descriptions & 12.1\% & 17.4\% \\
Fail to perform reasoning & 15.9\% & 13.7 \% \\
Select a less prominent element & 19.8\% & 24.8\% \\
Noisy annotation & 12.7\% & 10.5\% \\
\hline
\end{tabular}
\caption{
Error breakdowns of the embedding and alignment models
on 100 examples.
The embedding model handles implicit descriptions well, while
the alignment model excels at string matching.
}\label{tab:errors}
\end{table}

\paragraph{Fail to match strings.}
Many commands simply specify the text content of the element
(e.g., ``click customised garages'' $\to$ the link with text
``Customised Garages, Canopies \& Carports'').
The encoding model, which encodes the whole command as a single vector,
occasionally fails to select the element with partially matching texts.
In contrast, the alignment model explicitly models text matching,
and thus is better at this type of commands.

\paragraph{Incorrectly match strings.}
Due to its reliance on text matching,
the alignment model struggles when many elements
share substrings with the command
(e.g., ``shop for knitwear'' when many elements contain the word ``shop''),
or when an element with a matching substring is not the correct target
(e.g., ``get the program'' $\to$ the ``Download'' link,
not the ``Microsoft developer program'' link).

\paragraph{Fail to understand descriptions.}
As seen in \reftab{analysis},
many commands indirectly describe the elements using paraphrases,
goal descriptions, or properties of the elements.
The encoding model, which summarizes various properties of the elements
as an embedding vector,
is better at handling these commands
than the alignment model, but still makes a few errors on harder examples
(e.g., ``please close this notice for me'' $\to$ the ``X'' button with hidden text ``Hide``).

\paragraph{Fail to perform reasoning.}
For the most part, the models fail to handle relational, ordinal, or spatial reasoning.
The most frequent error mode is when the element is a text box, %
and the command uses nearby label as the reference.
While a few text boxes have semantic annotations
which the model can use
(e.g., tooltip or aria attributes),
many web pages do not provide such annotations.
To handle these cases, a model would have to identify the label
of the text box based on logical or visual contexts.

\paragraph{Other errors.}
Apart from the annotation noise,
occasionally multiple elements on the web page satisfy the given command
(e.g., ``log in'' can match any ``Sign In'' button on the web page).
In these cases, the annotation usually gives the most \emph{prominent} element
among the possible candidates.
To provide a natural interface for users,
the model should arguably learn to predict such prominent elements
instead of more obscure ones.
 
\section{Related work and discussion}

\paragraph{Mapping natural language to actions.}
Previous work on semantic parsing learns to perform actions
described by natural language utterances in various environments.
Examples of such actions include API calls 
\cite{young2013pomdp,su2017building,bordes2017learning},
database queries
\cite{zelle96geoquery,zettlemoyer07relaxed,berant2013freebase,yih2015stagg},
navigation
\cite{artzi2013weakly,janner2018representation},
and object manipulation
\cite{tellex2011understanding,andreas2015alignment,guu2017bridging,fried2018unified}.

For web pages and graphical user interfaces,
there are previous works on using natural language to
perform computations on web tables
\cite{pasupat2015compositional,zhong2017seq2sql}
and submit web forms
\cite{shi2017wob}.
Our task is similar to previous works on
interpreting instructions on user interfaces
\cite{branavan09reinforcement,branavan10high,liu2018workflow}.
While their works focuses on learning from distant supervision,
we consider shallower interactions but on a much broader domain.

Previous work also explores the reverse problem of generating
natural language description of objects
\cite{vinyals2014show,karpathy2015deep,zarriass2017obtaining}.
We hope that our dataset could also be useful
for exploring the reverse task of describing actions on web pages.

\paragraph{Reference games.}
In a reference game,
the system has to select the correct object referenced by the given utterance
\cite{frank2012pragmatics}.
Previous work on reference games
focuses on a small number of objects with similar properties,
and applies pragmatics to handle ambiguous utterance
\cite{golland2010pragmatics,smith2013pragmatics,celikyilmaz2014resolving,andreas2016reasoning,yu2017joint}.
Our task can be viewed as a reference game with several challenges:
higher number of objects, diverse object properties,
and the need to interpret objects based on their contexts.

\paragraph{Interacting with web pages.}
Automated scripts are used to interact with web elements.
While most scripts reference elements with logical selectors (e.g., CSS and XPath),
there have been several alternatives such as images
\cite{yeh2009sikuli}
and simple natural language utterances
\cite{soh2017tagui}.
Some other interfaces for navigating web pages
include keystrokes \cite{spalteholz2008keysurf},
speech \cite{ashok2014wizard},
haptics \cite{yu2005haptics},
and eye gaze \cite{kumar2007eyepoint}.

\section{Conclusion}
We presented a new task
of grounding natural language commands on
open-ended and semi-structured
web pages.
With different methods of referencing elements,
mixtures of textual and non-textual element attributes,
and the need to properly incorporate context,
our task offers a challenging environment
for language understanding
with great potential for real-world applications.

Our dataset and code are available at
{\small\url{https://github.com/stanfordnlp/phrasenode}}.\\
Reproducible experiments are available on the CodaLab platform at
{\small\url{https://worksheets.codalab.org/worksheets/0x0097f249cd944284a81af331093c3579/}}.

\section*{Acknowledgments}
This work was supported by
NSF CAREER Award under No. IIS-1552635
and an Amazon Research Award.
 
\bibliography{all}

\begin{thebibliography}{46}
\expandafter\ifx\csname natexlab\endcsname\relax\def\natexlab#1{#1}\fi

\bibitem[{Andreas and Klein(2015)}]{andreas2015alignment}
J.~Andreas and D.~Klein. 2015.
\newblock Alignment-based compositional semantics for instruction following.
\newblock In \emph{Empirical Methods in Natural Language Processing (EMNLP)}.

\bibitem[{Andreas and Klein(2016)}]{andreas2016reasoning}
J.~Andreas and D.~Klein. 2016.
\newblock Reasoning about pragmatics with neural listeners and speakers.
\newblock In \emph{Empirical Methods in Natural Language Processing (EMNLP)},
  pages 1173--1182.

\bibitem[{Artzi and Zettlemoyer(2013)}]{artzi2013weakly}
Y.~Artzi and L.~Zettlemoyer. 2013.
\newblock Weakly supervised learning of semantic parsers for mapping
  instructions to actions.
\newblock \emph{Transactions of the Association for Computational Linguistics
  (TACL)}, 1:49--62.

\bibitem[{Ashok et~al.(2014)Ashok, Borodin, Stoyanchev, Puzis, and
  Ramakrishnan}]{ashok2014wizard}
V.~Ashok, Y.~Borodin, S.~Stoyanchev, Y.~Puzis, and I.~V. Ramakrishnan. 2014.
\newblock Wizard-of-{O}z evaluation of speech-driven web browsing interface for
  people with vision impairments.
\newblock In \emph{Web for All Conference}.

\bibitem[{Berant et~al.(2013)Berant, Chou, Frostig, and
  Liang}]{berant2013freebase}
J.~Berant, A.~Chou, R.~Frostig, and P.~Liang. 2013.
\newblock Semantic parsing on {F}reebase from question-answer pairs.
\newblock In \emph{Empirical Methods in Natural Language Processing (EMNLP)}.

\bibitem[{Bordes and Weston(2017)}]{bordes2017learning}
A.~Bordes and J.~Weston. 2017.
\newblock Learning end-to-end goal-oriented dialog.
\newblock In \emph{International Conference on Learning Representations
  (ICLR)}.

\bibitem[{Branavan et~al.(2009)Branavan, Chen, Zettlemoyer, and
  Barzilay}]{branavan09reinforcement}
S.~Branavan, H.~Chen, L.~S. Zettlemoyer, and R.~Barzilay. 2009.
\newblock Reinforcement learning for mapping instructions to actions.
\newblock In \emph{Association for Computational Linguistics and International
  Joint Conference on Natural Language Processing (ACL-IJCNLP)}, pages 82--90.

\bibitem[{Branavan et~al.(2010)Branavan, Zettlemoyer, and
  Barzilay}]{branavan10high}
S.~Branavan, L.~Zettlemoyer, and R.~Barzilay. 2010.
\newblock Reading between the lines: Learning to map high-level instructions to
  commands.
\newblock In \emph{Association for Computational Linguistics (ACL)}, pages
  1268--1277.

\bibitem[{\c{C}elikyilmaz et~al.(2014)\c{C}elikyilmaz, Feizollahi,
  Hakkani-T{\"u}r, and Sarikaya}]{celikyilmaz2014resolving}
A.~\c{C}elikyilmaz, Z.~Feizollahi, D.~Z. Hakkani-T{\"u}r, and R.~Sarikaya.
  2014.
\newblock Resolving referring expressions in conversational dialogs for natural
  user interfaces.
\newblock In \emph{Empirical Methods in Natural Language Processing (EMNLP)}.

\bibitem[{Chen and Mooney(2011)}]{chen11navigate}
D.~L. Chen and R.~J. Mooney. 2011.
\newblock Learning to interpret natural language navigation instructions from
  observations.
\newblock In \emph{Association for the Advancement of Artificial Intelligence
  (AAAI)}, pages 859--865.

\bibitem[{Frank and Goodman(2012)}]{frank2012pragmatics}
M.~Frank and N.~D. Goodman. 2012.
\newblock Predicting pragmatic reasoning in language games.
\newblock \emph{Science}, 336:998--998.

\bibitem[{Fried et~al.(2018)Fried, Andreas, and Klein}]{fried2018unified}
D.~Fried, J.~Andreas, and D.~Klein. 2018.
\newblock Unified pragmatic models for generating and following instructions.
\newblock In \emph{North American Association for Computational Linguistics
  (NAACL)}.

\bibitem[{Golland et~al.(2010)Golland, Liang, and
  Klein}]{golland2010pragmatics}
D.~Golland, P.~Liang, and D.~Klein. 2010.
\newblock A game-theoretic approach to generating spatial descriptions.
\newblock In \emph{Empirical Methods in Natural Language Processing (EMNLP)},
  pages 410--419.

\bibitem[{Guu et~al.(2017)Guu, Pasupat, Liu, and Liang}]{guu2017bridging}
K.~Guu, P.~Pasupat, E.~Z. Liu, and P.~Liang. 2017.
\newblock From language to programs: Bridging reinforcement learning and
  maximum marginal likelihood.
\newblock In \emph{Association for Computational Linguistics (ACL)}.

\bibitem[{Hammoudi et~al.(2016)Hammoudi, Rothermel, and
  Tonella}]{hammoudi2016why}
M.~Hammoudi, G.~Rothermel, and P.~Tonella. 2016.
\newblock Why do record/replay tests of web applications break?
\newblock \emph{IEEE International Conference on Software Testing, Verification
  and Validation}.

\bibitem[{Hu et~al.(2014)Hu, Lu, Li, and Chen}]{hu2014convolutional}
B.~Hu, Z.~Lu, H.~Li, and Q.~Chen. 2014.
\newblock Convolutional neural network architectures for matching natural
  language sentences.
\newblock In \emph{Advances in Neural Information Processing Systems (NIPS)}.

\bibitem[{Janner et~al.(2018)Janner, Narasimhan, and
  Barzilay}]{janner2018representation}
M.~Janner, K.~Narasimhan, and R.~Barzilay. 2018.
\newblock Representation learning for grounded spatial reasoning.
\newblock \emph{Transactions of the Association for Computational Linguistics
  (TACL)}, 6.

\bibitem[{Karpathy and Fei-Fei(2015)}]{karpathy2015deep}
A.~Karpathy and L.~Fei-Fei. 2015.
\newblock Deep visual-semantic alignments for generating image descriptions.
\newblock In \emph{Computer Vision and Pattern Recognition (CVPR)}, pages
  3128--3137.

\bibitem[{Kingma and Ba(2014)}]{kingma2014adam}
D.~Kingma and J.~Ba. 2014.
\newblock Adam: A method for stochastic optimization.
\newblock \emph{arXiv preprint arXiv:1412.6980}.

\bibitem[{Kiros et~al.(2015)Kiros, Zhu, Salakhutdinov, Zemel, Urtasun,
  Torralba, and Fidler}]{kiros2015skip}
R.~Kiros, Y.~Zhu, R.~Salakhutdinov, R.~S. Zemel, R.~Urtasun, A.~Torralba, and
  S.~Fidler. 2015.
\newblock Skip-thought vectors.
\newblock In \emph{Advances in Neural Information Processing Systems (NIPS)}.

\bibitem[{Kumar et~al.(2007)Kumar, Paepcke, and Winograd}]{kumar2007eyepoint}
M.~Kumar, A.~Paepcke, and T.~Winograd. 2007.
\newblock Eyepoint: practical pointing and selection using gaze and keyboard.
\newblock In \emph{Conference on Human Factors in Computing Systems (CHI)}.

\bibitem[{Liu et~al.(2018)Liu, Guu, Pasupat, Shi, and Liang}]{liu2018workflow}
E.~Z. Liu, K.~Guu, P.~Pasupat, T.~Shi, and P.~Liang. 2018.
\newblock Reinforcement learning on web interfaces using workflow-guided
  exploration.
\newblock In \emph{International Conference on Learning Representations
  (ICLR)}.

\bibitem[{Misra et~al.(2015)Misra, Tao, Liang, and
  Saxena}]{misra2015environment}
D.~K. Misra, K.~Tao, P.~Liang, and A.~Saxena. 2015.
\newblock Environment-driven lexicon induction for high-level instructions.
\newblock In \emph{Association for Computational Linguistics (ACL)}.

\bibitem[{Pasupat and Liang(2015)}]{pasupat2015compositional}
P.~Pasupat and P.~Liang. 2015.
\newblock Compositional semantic parsing on semi-structured tables.
\newblock In \emph{Association for Computational Linguistics (ACL)}.

\bibitem[{Seo et~al.(2016)Seo, Kembhavi, Farhadi, and
  Hajishirzi}]{seo2016bidaf}
M.~Seo, A.~Kembhavi, A.~Farhadi, and H.~Hajishirzi. 2016.
\newblock Bidirectional attention flow for machine comprehension.
\newblock \emph{arXiv}.

\bibitem[{Shi et~al.(2017)Shi, Karpathy, Fan, Hernandez, and
  Liang}]{shi2017wob}
T.~Shi, A.~Karpathy, L.~Fan, J.~Hernandez, and P.~Liang. 2017.
\newblock World of bits: An open-domain platform for web-based agents.
\newblock In \emph{International Conference on Machine Learning (ICML)}.

\bibitem[{Smith et~al.(2013)Smith, Goodman, and Frank}]{smith2013pragmatics}
N.~J. Smith, N.~D. Goodman, and M.~C. Frank. 2013.
\newblock Learning and using language via recursive pragmatic reasoning about
  other agents.
\newblock In \emph{Advances in Neural Information Processing Systems (NIPS)},
  pages 3039--3047.

\bibitem[{Soh(2017)}]{soh2017tagui}
K.~Soh. 2017.
\newblock {TagUI}: {RPA} / {CLI} tool for automating user interactions.
\newblock \url{https://github.com/kelaberetiv/TagUI}.

\bibitem[{Spalteholz et~al.(2008)Spalteholz, Li, Livingston, and
  Hamidi}]{spalteholz2008keysurf}
L.~Spalteholz, K.~F. Li, N.~Livingston, and F.~Hamidi. 2008.
\newblock Keysurf: a character controlled browser for people with physical
  disabilities.
\newblock In \emph{World Wide Web (WWW)}.

\bibitem[{Su et~al.(2017)Su, Awadallah, Khabsa, Pantel, Gamon, and
  Encarnaci\'{o}n}]{su2017building}
Y.~Su, A.~H. Awadallah, M.~Khabsa, P.~Pantel, M.~Gamon, and M.~J.
  Encarnaci\'{o}n. 2017.
\newblock Building natural language interfaces to web apis.
\newblock In \emph{Conference on Information and Knowledge Management (CIKM)}.

\bibitem[{Tai et~al.(2015)Tai, Socher, and Manning}]{tai2015improved}
K.~S. Tai, R.~Socher, and C.~D. Manning. 2015.
\newblock Improved semantic representations from tree-structured long
  short-term memory networks.
\newblock In \emph{Association for Computational Linguistics (ACL)}.

\bibitem[{Tellex et~al.(2011)Tellex, Kollar, Dickerson, Walter, Banerjee,
  Teller, and Roy}]{tellex2011understanding}
S.~Tellex, T.~Kollar, S.~Dickerson, M.~R. Walter, A.~G. Banerjee, S.~J. Teller,
  and N.~Roy. 2011.
\newblock Understanding natural language commands for robotic navigation and
  mobile manipulation.
\newblock In \emph{Association for the Advancement of Artificial Intelligence
  (AAAI)}.

\bibitem[{Vinyals et~al.(2014)Vinyals, Toshev, Bengio, and
  Erhan}]{vinyals2014show}
O.~Vinyals, A.~Toshev, S.~Bengio, and D.~Erhan. 2014.
\newblock Show and tell: A neural image caption generator.
\newblock \emph{arXiv preprint arXiv:1411.4555}.

\bibitem[{Xiong et~al.(2017)Xiong, Zhong, and Socher}]{xiong2017dynamic}
C.~Xiong, V.~Zhong, and R.~Socher. 2017.
\newblock Dynamic coattention networks for question answering.
\newblock In \emph{International Conference on Learning Representations
  (ICLR)}.

\bibitem[{Yeh et~al.(2009)Yeh, Chang, and Miller}]{yeh2009sikuli}
T.~Yeh, T.~Chang, and R.~Miller. 2009.
\newblock Sikuli: using {GUI} screenshots for search and automation.
\newblock In \emph{User Interface Software and Technology (UIST)}.

\bibitem[{Yih et~al.(2015)Yih, Chang, He, and Gao}]{yih2015stagg}
W.~Yih, M.~Chang, X.~He, and J.~Gao. 2015.
\newblock Semantic parsing via staged query graph generation: Question
  answering with knowledge base.
\newblock In \emph{Association for Computational Linguistics (ACL)}.

\bibitem[{Yin et~al.(2016)Yin, Sch{\"u}tze, Xiang, and Zhou}]{yin2016abcnn}
W.~Yin, H.~Sch{\"u}tze, B.~Xiang, and B.~Zhou. 2016.
\newblock {ABCNN}: Attention-based convolutional neural network for modeling
  sentence pairs.
\newblock \emph{Transactions of the Association for Computational Linguistics
  (TACL)}, 4.

\bibitem[{Young et~al.(2013)Young, Ga{\v{s}}i{\'c}, Thomson, and
  Williams}]{young2013pomdp}
S.~Young, M.~Ga{\v{s}}i{\'c}, B.~Thomson, and J.~D. Williams. 2013.
\newblock {POMDP}-based statistical spoken dialog systems: A review.
\newblock In \emph{Proceedings of the IEEE}, 5, pages 1160--1179.

\bibitem[{Yu et~al.(2018)Yu, Dohan, Luong, Zhao, Chen, Norouzi, and
  Le}]{yu2018qanet}
A.~W. Yu, D.~Dohan, M.~Luong, R.~Zhao, K.~Chen, M.~Norouzi, and Q.~V. Le. 2018.
\newblock {QANet}: Combining local convolution with global self-attention for
  reading comprehension.
\newblock In \emph{International Conference on Learning Representations
  (ICLR)}.

\bibitem[{Yu et~al.(2017)Yu, Tan, Bansal, and Berg}]{yu2017joint}
L.~Yu, H.~Tan, M.~Bansal, and T.~L. Berg. 2017.
\newblock A joint speaker-listener-reinforcer model for referring expressions.
\newblock In \emph{Computer Vision and Pattern Recognition (CVPR)}.

\bibitem[{Yu et~al.(2005)Yu, Kuber, Murphy, Strain, and
  McAllister}]{yu2005haptics}
W.~Yu, R.~Kuber, E.~Murphy, P.~Strain, and G.~McAllister. 2005.
\newblock A novel multimodal interface for improving visually impaired people's
  web accessibility.
\newblock \emph{Virtual Reality}, 9.

\bibitem[{Zajicek et~al.(1998)Zajicek, Powell, and Reeves}]{zajicek1998blind}
M.~Zajicek, C.~Powell, and C.~Reeves. 1998.
\newblock A web navigation tool for the blind.
\newblock In \emph{International ACM Conference on Assistive Technologies}.

\bibitem[{Zarriai{\ss} and Schlangen(2017)}]{zarriass2017obtaining}
S.~Zarriai{\ss} and D.~Schlangen. 2017.
\newblock Obtaining referential word meanings from visual and distributional
  information: Experiments on object naming.
\newblock In \emph{Association for Computational Linguistics (ACL)}.

\bibitem[{Zelle and Mooney(1996)}]{zelle96geoquery}
M.~Zelle and R.~J. Mooney. 1996.
\newblock Learning to parse database queries using inductive logic programming.
\newblock In \emph{Association for the Advancement of Artificial Intelligence
  (AAAI)}, pages 1050--1055.

\bibitem[{Zettlemoyer and Collins(2007)}]{zettlemoyer07relaxed}
L.~S. Zettlemoyer and M.~Collins. 2007.
\newblock Online learning of relaxed {CCG} grammars for parsing to logical
  form.
\newblock In \emph{Empirical Methods in Natural Language Processing and
  Computational Natural Language Learning (EMNLP/CoNLL)}, pages 678--687.

\bibitem[{Zhong et~al.(2017)Zhong, Xiong, and Socher}]{zhong2017seq2sql}
V.~Zhong, C.~Xiong, and R.~Socher. 2017.
\newblock Seq2sql: Generating structured queries from natural language using
  reinforcement learning.
\newblock \emph{arXiv preprint arXiv:1709.00103}.

\end{thebibliography}
\bibliographystyle{acl_natbib_nourl}

\appendix
 
\end{document}